\documentclass[10pt,twocolumn,letterpaper]{article}

\usepackage[pagenumbers]{cvpr} 

\usepackage{graphicx}
\usepackage{amsmath}
\usepackage{amssymb}
\usepackage{booktabs}
\usepackage[accsupp]{axessibility}

\usepackage[pagebackref,breaklinks,colorlinks]{hyperref}

\usepackage[capitalize]{cleveref}
\crefname{section}{Sec.}{Secs.}
\Crefname{section}{Section}{Sections}
\Crefname{table}{Table}{Tables}
\crefname{table}{Tab.}{Tabs.}

\def\cvprPaperID{18} 
\def\confName{CVPR}
\def\confYear{2023}

\begin{document}

\title{Making the V in Text-VQA Matter}

\author{Shamanthak Hegde\\
KLE Technological University\\
Hubballi, India\\
{\tt\small 01fe19bcs233@kletech.ac.in}
\and
Soumya Jahagirdar\\
CVIT, IIIT Hyderabad\\
Hyderabad, India\\
{\tt\small soumya.jahagirdar@research.iiit.ac.in}
\and
Shankar Gangisetty\\
IIIT Hyderabad\\
Hyderabad, India\\
{\tt\small shankar.gangisetty@ihub-data.iiit.ac.in}
}
\maketitle

\begin{abstract}
    Text-based VQA aims at answering questions by reading the text present in the images. It requires a large amount of scene-text relationship understanding compared to the VQA task. Recent studies have shown that the question-answer pairs in the dataset are more focused on the text present in the image but less importance is given to visual features and some questions do not require understanding the image. The models trained on this dataset predict biased answers due to the lack of understanding of visual context. For example, in questions like “What is written on the signboard?”, the answer predicted by the model is always “STOP” which makes the model to ignore the image. To address these issues, we propose a method to learn visual features (making V matter in TextVQA) along with the OCR features and question features using VQA dataset as external knowledge for Text-based VQA. Specifically, we combine the TextVQA dataset and VQA dataset and train the model on this combined dataset. Such a simple, yet effective approach increases the understanding and correlation between the image features and text present in the image, which helps in the better answering of questions. We further test the model on different datasets and compare their qualitative and quantitative results.
\end{abstract}


\section{Introduction}

\label{sec:intro}
\begin{figure}
    \centering
    \includegraphics[width=1\linewidth]{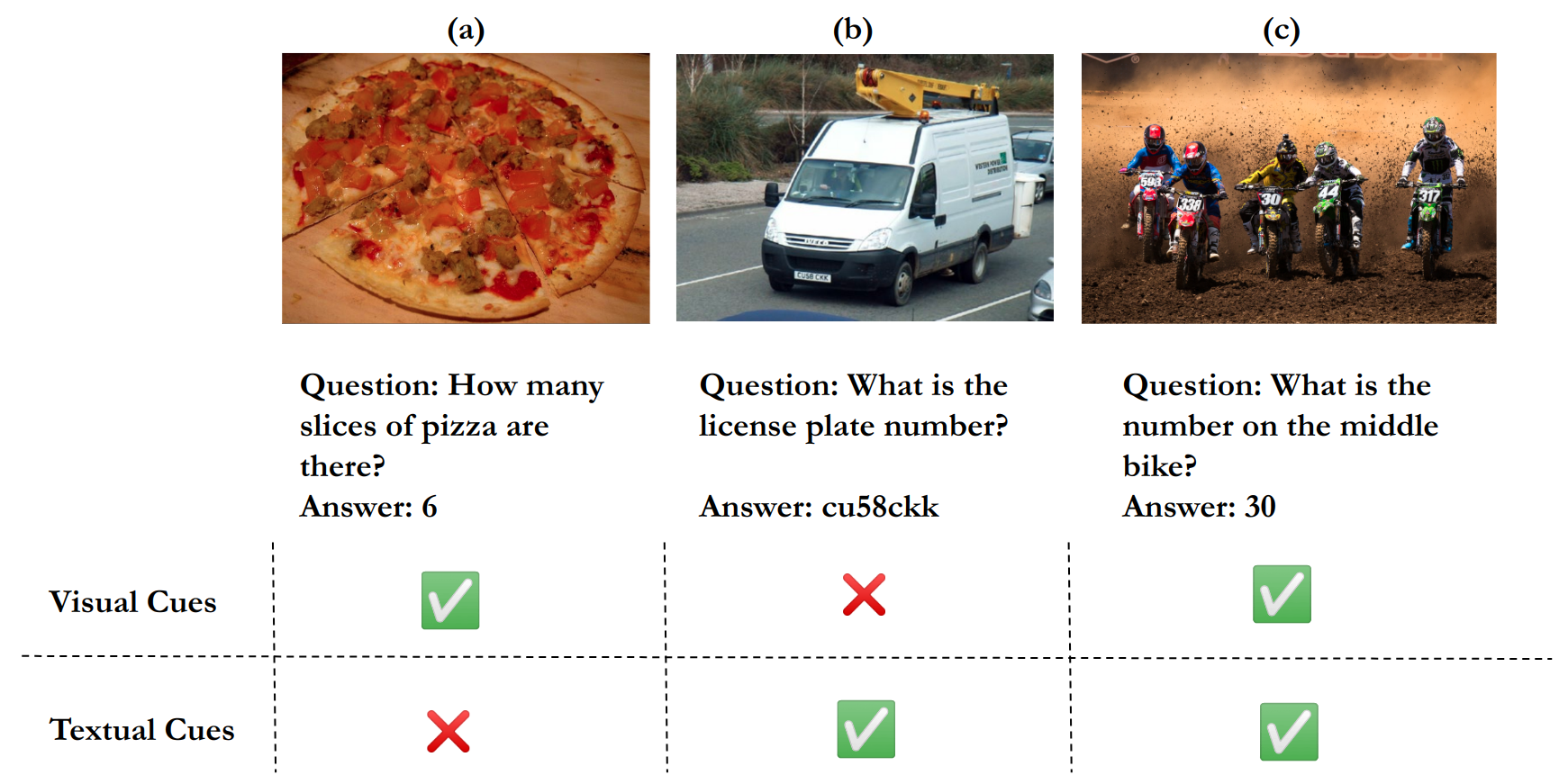}
    \caption{(a) Visual cues can provide additional context and help clarify meaning, (b) while textual cues can provide more detailed information to support understanding. (c) Combined cues that integrate both visual and textual information can be powerful. 
    } 
    \label{fig:main_fig}
\end{figure}

In recent years, deep learning models that require an understanding of visual scenes by answering questions about everyday scenes have become important. Towards this, many works~\cite{first-vqa:2017,vcr:2019}
have introduced datasets and methods that present varied types of questions over different scenes. A few works~\cite{textVQA,stvqa,ocr-vqa,docvqa:2021, textvqg} have focused on the datasets that require models to read the text present in the images. Recently a few works have introduced works on text-based video question answering \cite{video_text_videoqa, newsvideoqa}. These datasets and works provide methods with the ability to learn to answer questions belonging to a certain domain. In Fig.~\ref{fig:main_fig}, we can see that dataset are designed by keeping the domain into consideration. Question-answer pairs in the VQA dataset are framed based on the visual scene only. One such example is seen in Fig.~\ref{fig:main_fig}(a) \textit{``How many slices of pizza are there?''} . 

On other hand, datasets like TextVQA~\cite{textVQA} have questions majorly that require textual content in the image to answer the questions and little to no visual information is needed to obtain the answer. An example is shown in Fig.~\ref{fig:main_fig}(b) where the question is \textit{``What is the license plate number?''} and answers can be obtained just by using the OCR information. Ideally, a good VQA system should be able to \textbf{look} and \textbf{read} as shown in Fig.~\ref{fig:main_fig} (c) where the question is \textit{``What is the number on the middle of the bike?''}. To answer this question, the model should first look at the image and find the region ``middle'' as instructed by the question. Then, the model should read the number written on the region of interest. Current methods and datasets suffer from bias of domain-specific questions that result in these methods to learn shortcuts to obtain the answers. In~\cite{making_v_matter_vqa}, the authors introduce a technique of augmenting dataset to remove these language priors existing in the VQA dataset. Complementary images are added to the existing VQA dataset such that language priors are removed. On the other hand, authors of \cite{structured_multimodal_attention, towards_explainable_textvqa_model}, show that there is an obvious gap in the TextVQA models to learn to look at the images while answering the questions and VQA models to learn to read. Though the models trained specifically on domain specific dataset perform very well, they tend to fail when questions from other domains are asked. The language priors in these domain-specific datasets make the existing methods under exploit combining information from multiple modalities but only use these language priors to obtain higher accuracy. To subside this effect of language priors for the task of Text-based VQA, we propose a method to \textbf{Make the V in TextVQA Matter}.

In this work, to dissolve this bias, we propose a new method of multimodal training on the union of Visual Question Answering (VQA)~\cite{first-vqa:2017} and Text-based Visual Question Answering (TextVQA~\cite{textVQA} + ST-VQA~\cite{stvqa}) datasets. Specifically, we balance the Text-based VQA dataset by adding the images in the VQA dataset which contain text in them. We call this merged dataset as \textbf{Union Dataset}. Our dataset is more balanced compared to TextVQA only and VQA only in terms of types of questions that require both looking and reading the image to answer. We train the state-of-the-art models of TextVQA task on the union dataset and perform exhaustive experiments. These models include iterative answer prediction with pointer-augmented multimodal transformers for TextVQA \cite{M4C:2020} and text-aware pre-training for Text-VQA and Text-Caption \cite{TAP}. We provide attention maps for a better understanding of the proposed method and compare them with attention maps obtained from existing methods. We also show the generalization of such a method to new datasets like \cite{kvqa} by directly testing on the new dataset and also by fine-tuning on it.\\
Our main contributions are as follows: 
\begin{enumerate}
    \item We balance the current Text-based VQA datasets by combining (union) images from VQA dataset such that the images should contain textual information. This results in the dataset twice as only Text-based VQA dataset with questions that will make the methods learn to look and read.
    \item We evaluate state-of-the-art TextVQA models on the proposed union dataset and show that the models trained on existing out of balance datasets exploit the language prior to obtain answer. This observation helps our premise that combining the datasets can help in making the visual information matter in TextVQA. In addition to this, we test our hypothesis and show that it generalizes well on new out of domain test set. We show this by evaluating the performance of the model trained with our union dataset on KVQA \cite{kvqa} dataset.
\end{enumerate}

\begin{figure}
    \centering
    \includegraphics[width=0.85\linewidth]{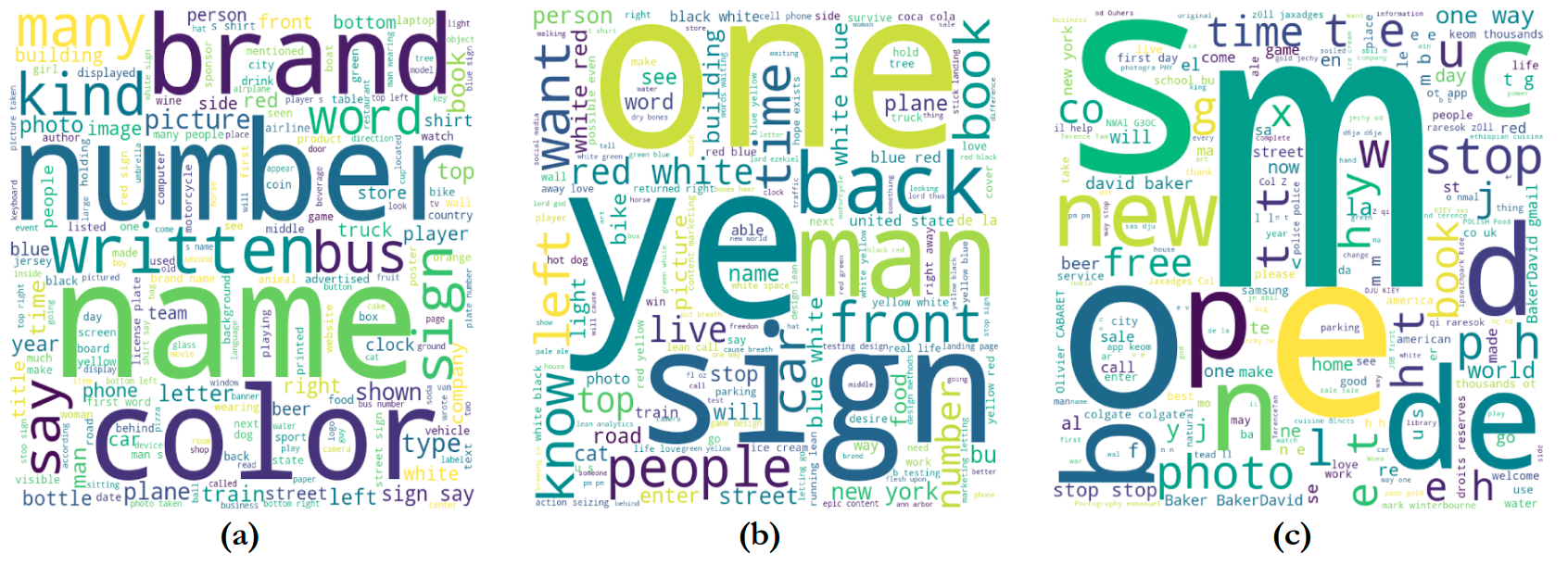}
    \caption{ (a) Wordcloud for words in question in our union dataset. (b) Wordcloud for words in answers. (c) Wordcloud for words in OCR tokens.
    } 
    \label{fig:wordcloud}
\end{figure}

\begin{figure*}
    \centering
    \includegraphics [width=0.85\linewidth, scale=3]{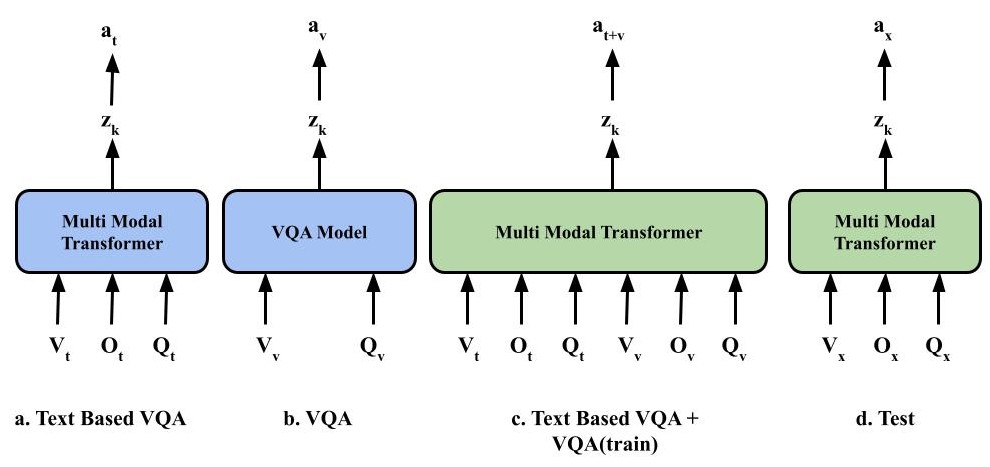}
    \caption{Different models used for TextVQA and VQA and combined tasks.(a)  The existing method for Text-based VQA using Multi-Modal Transformer. (b) Existing VQA models for VQA tasks. (c) Our method where we pass combined dataset of Text-based VQA and VQA datasets for training. (d) Testing our method on different datasets.}
    \label{fig:task_method}
\end{figure*}

\section{Related Work}
\label{sec:related}

\subsection{Debiasing in Visual Question Answering}
In VQA bias is ubiquitous, existing VQA~\cite{first-vqa:2017} dataset have biases between the questions and answers.  For example, (i) strong correlations between questions and answers, i.e., \textit{language prior}~\cite{AgrawalBPK18,making_v_matter_vqa} such as answering \textit{``green''} for the question \textit{``What color is the grass?''}, \textit{``tennis''} for the question \textit{``What sports ... ''} will obtain 40\% accuracy~\cite{counterfactual_vqa:2021}, (ii) questioner tends to ask about the objects seen in the image, i.e., \textit{visual priming bias}~\cite{making_v_matter_vqa, first-vqa:2017} such as answering \textit{``yes''} to all the questions \textit{``Do you see a ...''} achieves nearly 90\% accuracy because the model is trained and tested on the quite different scenarios. Recently, many methods have been proposed to overcome the biases in VQA. These methods can be classified as (i) non-augmentation-based methods~\cite{debiasing:2022, counterfactual_vqa:2021, lang_bias:2020, hint_bais:2019, rubi_bais:2019} seek to reduce the language biases explicitly or improve attention on the image (ii) augmentation-based methods~\cite{unshuffle_bias:2021, overcoming_bias:2020, mutant_bias:2020, counterfactual_vl:2020} seek to balance the biased dataset for unbiased training.

In \cite{debiasing:2022}, the authors use a dual masking strategy, wherein they train a VQA model by masking the most relevant image region or the question words and they use a negative answer assignment mechanism for providing the answers to the counterfactual samples synthesized which exploits the probability distribution of the answers based on their frequency in the original training set. In CF-VQA~\cite{counterfactual_vqa:2021}, the authors make use of both question and image, but use the two modalities individually without combining them. They subtract the pure language bias effect from the multimodal knowledge of standard VQA models. In \cite{overcoming_bias:2020}, the authors have proposed a self-supervised learning framework for VQA to automatically balance the biased data. They make use of an auxiliary task named question-image correlation estimation (QICE) to estimate the relevance between questions and images and generate a set of balanced question-image pairs with binary labels as either relevant or irrelevant, which are then used by the self-supervised auxiliary task to assist the VQA model to overcome language priors. In \cite{unshuffle_bias:2021}, the authors propose a general method to improve OOD generalization. The model is discouraged from using spurious correlations that only appear in subsets of the training data, and rather ensure that it uses reliable ones that are more likely to generalize at test time. More precisely, data is partitioned into multiple training environments such that spurious correlations vary across environments while reliable ones remain stable. by using unsupervised clustering, prior knowledge, and auxiliary annotations in existing datasets. Then, multiple copies of a neural network, one per environment are trained. Some of their weights are shared across environments, while others are subject to a variance regularizer in parameter space. This leads the model to extract features that are stable across environments since they are optimized to be predictive under a classifier common to all environments.

\subsection{Biases in Text-based Visual Question Answering}
In text-based VQA we expect a model to answer truthfully based on the visual evidence contained in the image, scene text and the correct intention of the question. Unfortunately, this is not always the case even for state-of-the-art methods. Instead of exploiting the image and scene text to find the correct answer, most models frequently rely on spurious correlations and follow the bias that naturally exists within the training data. This severely limits the generalization of Text-based VQA models in real-world scenarios, where the test distribution of facts (e.g., colors, counts, objects, position of objects, etc.) is often different from the training distribution. Few works such as \cite{logos, towards_explainable_textvqa_model} make use of an M4C~\cite{M4C:2020} like multimodal transformer while additionally having to train a separate decoder to ground the answer with bounding boxes in LOGOS~\cite{logos} and a segmentation network to output segmentation maps of the answer region in MTXNet~\cite{towards_explainable_textvqa_model}. These work provide a visual analysis on the region of interest of the model while answering the question. However, as per our knowledge, we are the first to propose solution for debiasing text-based VQA.

\begin{figure*}
    \centering
    \includegraphics[width=1\linewidth]{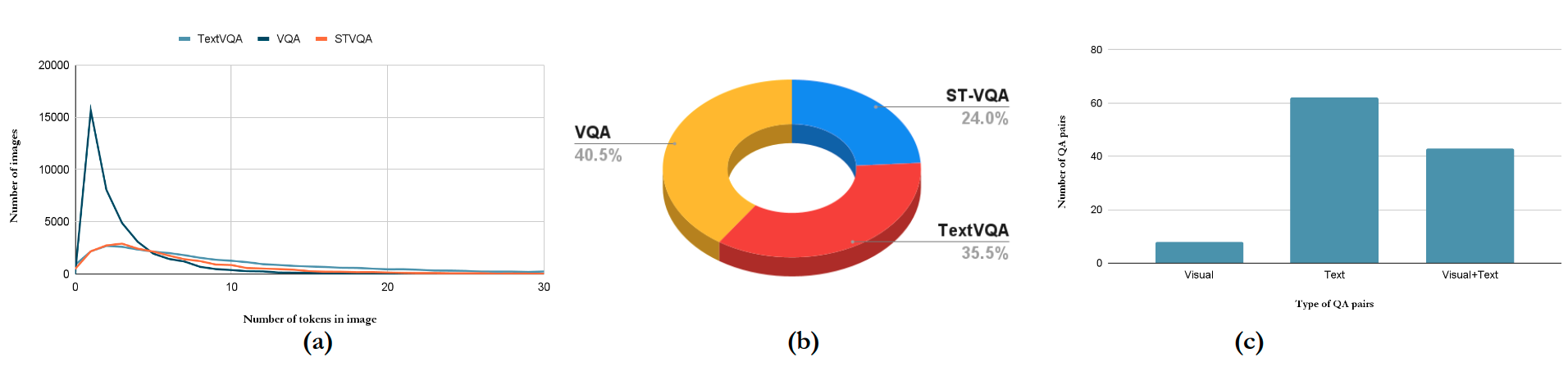}
    \caption{(a) Graph showing the distribution of length of OCR tokens in images of Union dataset. (b) Distribution of our Union Dataset. Our Union Dataset contains $35.5\%$ of question-answer pairs from TextVQA \cite{textVQA} dataset, $24.0\%$ question-answer pairs from ST-VQA \cite{stvqa} dataset, and $40.5\%$ question-answer pairs from VQA \cite{first-vqa:2017} dataset. (c) Bar chart showing ablation study when random of 100 QA pairs were given to human volunteers to classify each QA pair based on the answer based on Visual, Textual, or Visual+Textual.} 
    \label{fig:dist_dataset}
\end{figure*}

\section{Benchmarking Text-based VQA Models}
\label{lab:benchmarking}

In this section, we explain our method, as shown in Fig.~\ref{fig:task_method}. Fig.~\ref{fig:task_method}a and \ref{fig:task_method}b are models used for text-based VQA (TextVQA and ST-VQA) and VQA datasets. Then, we train a Multi Modal Transformer on the combined dataset wherein we specifically make use of two models M4C~\cite{M4C:2020} and TAP~\cite{TAP}. M4C~\cite{M4C:2020} is a multimodal transformer encoder with a dynamic pointer network decoder to select the answer from either vocabulary or detected OCR tokens. TAP~\cite{TAP} is an extended version of M4C which is pretrained on a large corpus of data, performing tasks such as masked language modelling(MLM), relative position prediction (RPP) and Image-Text Matching (ITM).

\subsection{Union of Visual and Text Based Datasets}
\label{subsec:union_dataset}

We combine Text-based VQA: TextVQA + ST-VQA, $Y = (V_t,O_t,Q_t)$ where $V_t$, $O_t$, $Q_t$ are objects, OCR, questions and VQA $Z = (V_v,O_v,Q_v)$ where $V_v$, $O_v$, $Q_v$ are objects, OCR, questions of VQA and call the combined dataset as \textbf{Union Dataset}. Fig.~\ref{fig:wordcloud} shows the word clouds for (a) words in the question of the Union dataset. (b) words in answers of Union dataset, and (c) words in OCR tokens of Union dataset. In Fig.~\ref{fig:dist_dataset} (a) shows the distribution of length of OCR tokens in images of Union dataset (TextVQA, VQA, STVQA). (b) Shows the \% distribution of TextVQA, ST-VQA and VQA dataset in Union dataset. It can be seen that, the Union dataset has balanced distribution.

We extract their corresponding object, OCR features along with the question feature to obtain dataset $W$. We consider the images from VQA \cite{first-vqa:2017} dataset that contains OCR tokens. Fig.~\ref{fig:dist_dataset}(b) shows the distribution of the number of question-answer pairs in each dataset. This union or merging of dataset results in balanced question-answer pairs which enable the Text-based VQA models to look (question-answer pairs from VQA dataset) and read (question-answer pairs from TextVQA and ST-VQA dataset). This results in current state-of-the-art Text-based VQA methods to learn to look and read.
\begin{equation}
    W = Y \cup Z
\end{equation}

\subsection{Multi-modal Transformer}
We use a multimodal transformer architecture containing three modalities – objects detected V, OCR tokens O and question words Q. We pass the feature embeddings to the model by projecting them in d-dimensional common embedding space with the following steps: 

\vspace{3pt}
\noindent
\textbf{Embedding of detected objects.} Given image I, we obtain N visual objects V (generally N is 100) and their corresponding location using a pretrained object detector (MaskRCNN). We consider the location of the $j^{th}$ object where $j=1,2,...,N$ by obtaining the relative bounding box $x^{b}_j$. We combine object feature $x^{fr}_j$ and bounding box $x^b_j$ to get the final object embedding $x^{obj}_j$ of the corresponding object $V_j$.

\begin{equation}
    {x^{obj}_j}={x^{fr}_j}+{x^b_j}
\end{equation}

\vspace{3pt}
\noindent
\textbf{OCR embedding.} We consider the M OCR tokens O (generally M is 50) extracted using EasyOCR (for VQA images) and Rosetta~\cite{rosetta} (for TextVQA and STVQA images). We extract the FastText word embedding feature $x^{ft}_i$ of the $i^{th}$ OCR token where $(i=1,2,...,M)$ along with the appearance feature $x^{ap}_i$, and the bounding box of each token, we sum it to get the OCR embedding $x^{ocr}_i$ of corresponding OCR token $O_i$.
\begin{equation}
    {x^{ocr}_i}={x^{ft}_i}+{x^b_i}
\end{equation}

\vspace{3pt}
\noindent
\textbf{Question words embedding.} We embed the Q question words (generally Q is 20) to a feature vector $x^{ques}$ using a pretrained BERT. We only use the first three layers of BERT to extract features of question words. \\ After embedding all entities from each modality as vectors in the d-dimensional joint embedding space, we apply a stack of L transformer layers\cite{attentionis_ayn} with a hidden dimension of d over the list of all entities. Through the multihead self-attention mechanism in transformers, each entity is allowed to freely attend to all other entities. Using the same transformer layers as a decoder, we predict the answer word by word in an autoregressive manner for a total of T steps.

\begin{figure*}
    \centering
    \includegraphics[width=.9\linewidth]{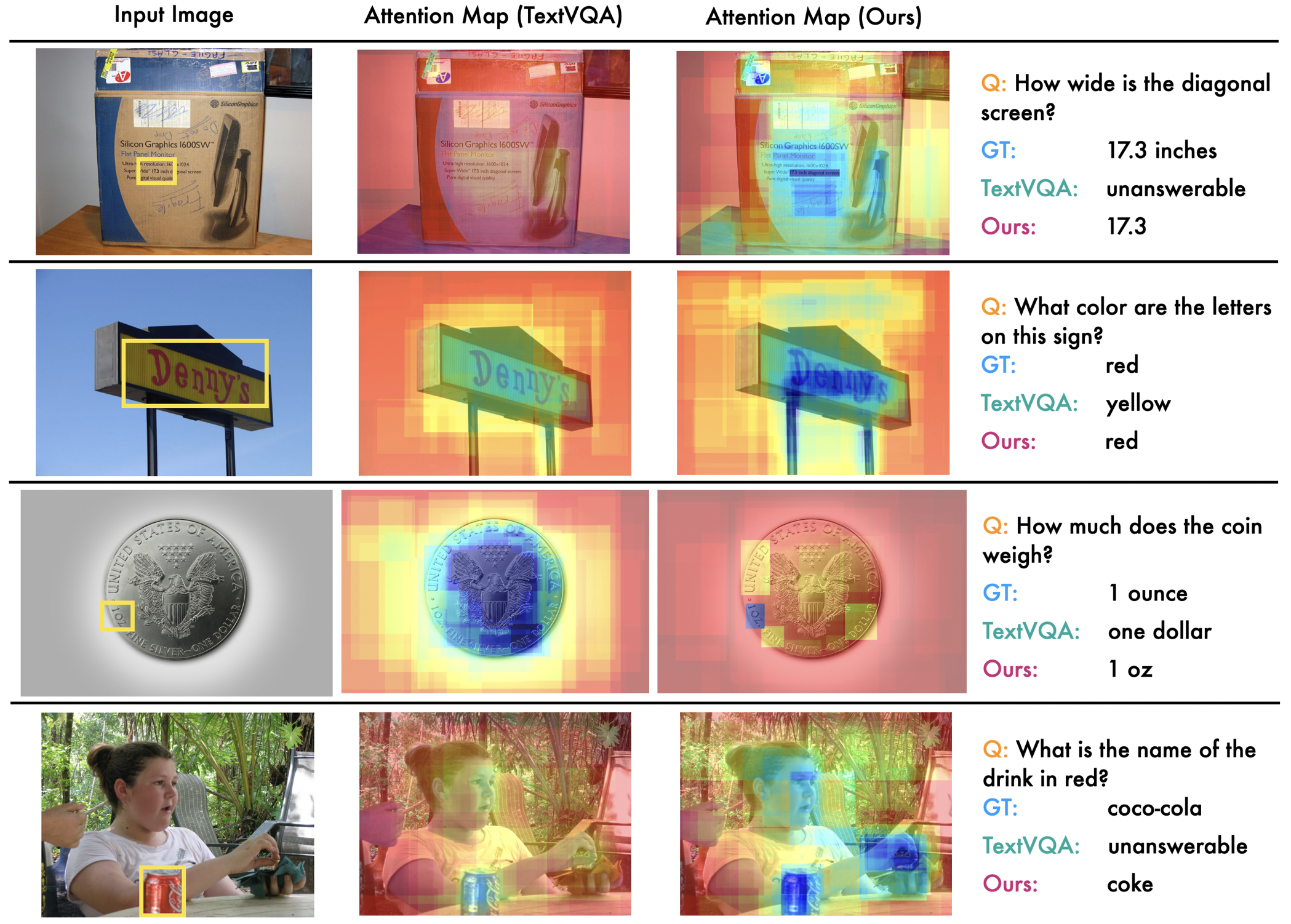}
    \caption{Qualitative results of TAP: Comparison on TextVQA and Ours with attention maps.}
    \label{fig:qualitative_tap}
\end{figure*}

\begin{table*}[hbt]
\centering
\begin{tabular}{l c c r}
\toprule
{\textbf{Method}} & {\textbf{Data for pretraining}}                                       & \textbf{Data for Finetuning data} & \textbf{Test Acc.} \\ \midrule
M4C             & -                                                                & TextVQA                  & 39.01                  \\ 
M4C             & -                                                                & \begin{tabular}[c]{@{}c@{}}TextVQA + VQA + STVQA\end{tabular}    & 39.16              \\
TAP             & TextVQA                                                          & -                        & 49.71              \\ 
TAP             & \begin{tabular}[c]{@{}c@{}}TextVQA + VQA + STVQA\end{tabular} & TextVQA                  & 47.75       \\  \bottomrule        
\end{tabular}
\caption{Evaluation of TextVQA \cite{textVQA} test data on text-based VQA models trained on our Union dataset. It can be seen that combining data from multiple sources helps the models that can only read to also look at the images thereby answering questions that require both textual and visual reasoning.}
\label{tab:textvqa_quant}
\end{table*}

\vspace{3pt}
\section{Experiments}
\label{sec:results}

In this section, we experiment and validate the performance of the proposed method of combining VQA and Text-based VQA datasets for better generalization of VQA systems that can see, read and reason. We first discuss the datasets in Sec.~\ref{subsec:datasets}. The quantitative and qualitative results are presented in Sec.~\ref{subsec:quan_res} and \ref{subsec:qual_res} respectively. We also provide several ablation studies in Sec.~\ref{subsec:ablation}.

\subsection{Datasets}
\label{subsec:datasets}
To showcase the effectiveness of the proposed method, we make use of three popular datasets, namely, VQA~\cite{first-vqa:2017}, TextVQA~\cite{textVQA} and ST-VQA~\cite{stvqa}. We obtain the images in the VQA dataset that contains text and combine them with {TextVQA+STVQA} dataset to obtain a union dataset: \textbf{VQA+TextVQA+STVQA}. We organize the test set of TextVQA as a test dataset for the models trained on the union dataset. We also evaluate our method on KVQA~\cite{kvqa} to show the generalization and domain transfer of the knowledge learned from the union dataset to a specific domain dataset such as KVQA. The training set of union datasets comprises 97,578 question-answer pairs and the test set contains 5,734 question-answer pairs.

\begin{figure*}[!htb]
    \centering
    \includegraphics[width=.9\linewidth]{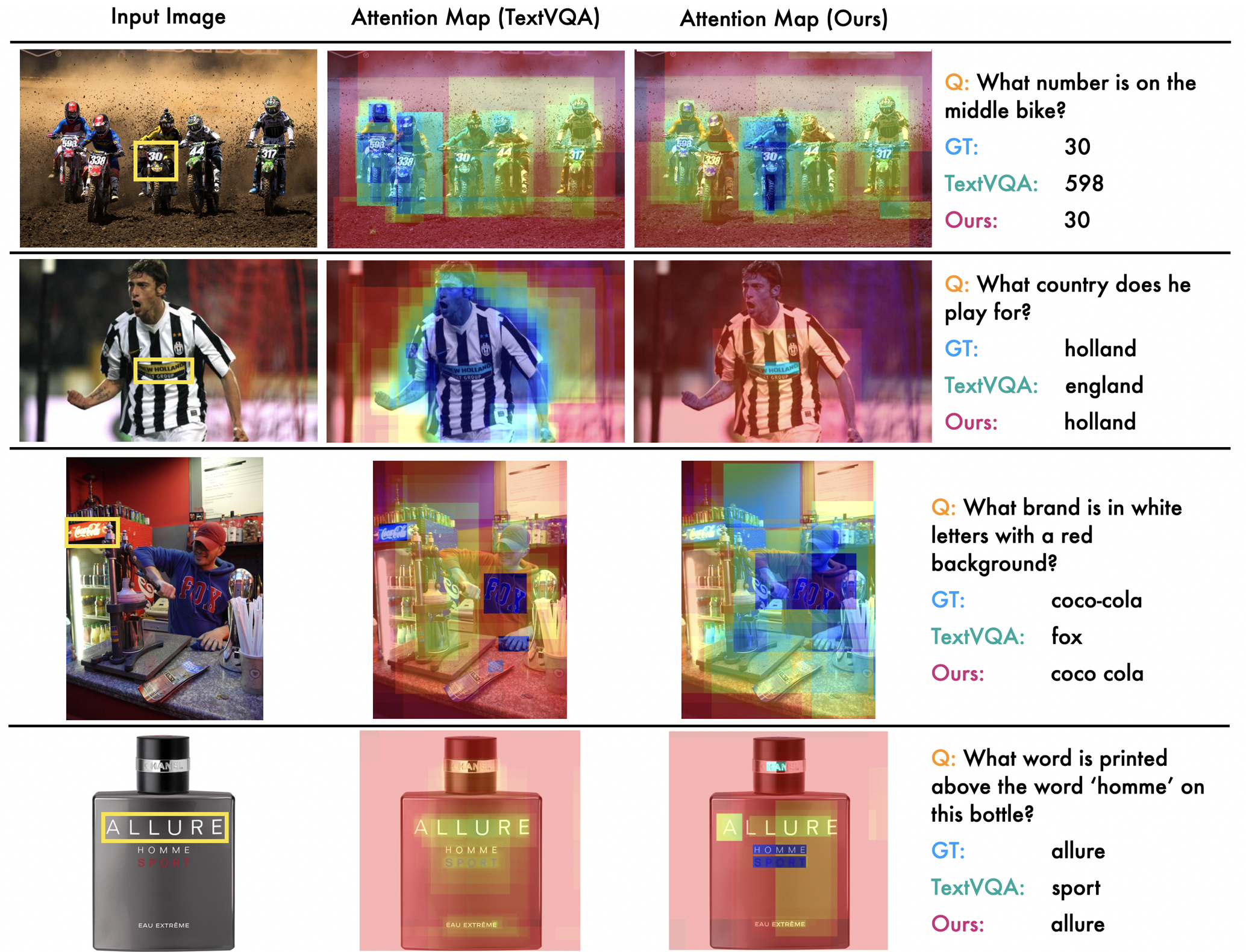}
    \caption{Qualitative results of M4C: Comparison on TextVQA and Ours with attention maps. It can be seen that M4C trained on \textbf{Union dataset} performs better for the questions that require models that need both visual and textual explanations to answer questions.}
    \label{fig:qualitative_m4c}
\end{figure*}

\noindent
\subsection{Performance metrics.}
We use Accuracy as the evaluation metric as it measures the percentage of questions for which the predicted answer matches exactly with atleast three of the target answers for the question. 
\begin{equation}
    Acc(\text{ans}) = min\left(\frac{\text{No of humans that said ans}}{3},1\right)
\end{equation}
We also show attention maps obtained from M4C~\cite{M4C:2020} and TAP~\cite{TAP} trained with the original configurations and trained on the union dataset.

\noindent
\subsection{Implementation details.}
The proposed method can be applied to different and newer approaches proposed in Visual Question Answering. The key idea is, making the visual cues important in text-based visual question answering. We used AdamW as the optimizer. The learning rate for the union dataset is set to $1e-4$. We train M4C \cite{M4C:2020} and TAP \cite{TAP} for 24,000 iterations with a batch size of 64. We recognize a maximum of 50 OCR tokens in the union dataset and detect a maximum of 100 objects from the image. We set the maximum decoding steps to 12 and use the answer vocabulary from the union dataset. In the case of experiments on test set of TextVQA, we train M4C on union dataset (TextVQA+STVQA+VQA) and directly evaluate on test set of TextVQA. To evaluate the generalization of the proposed method, we evaluate it on test set of KVQA.

\begin{figure*}[!htb]
    \centering
    \includegraphics[width=1\linewidth]{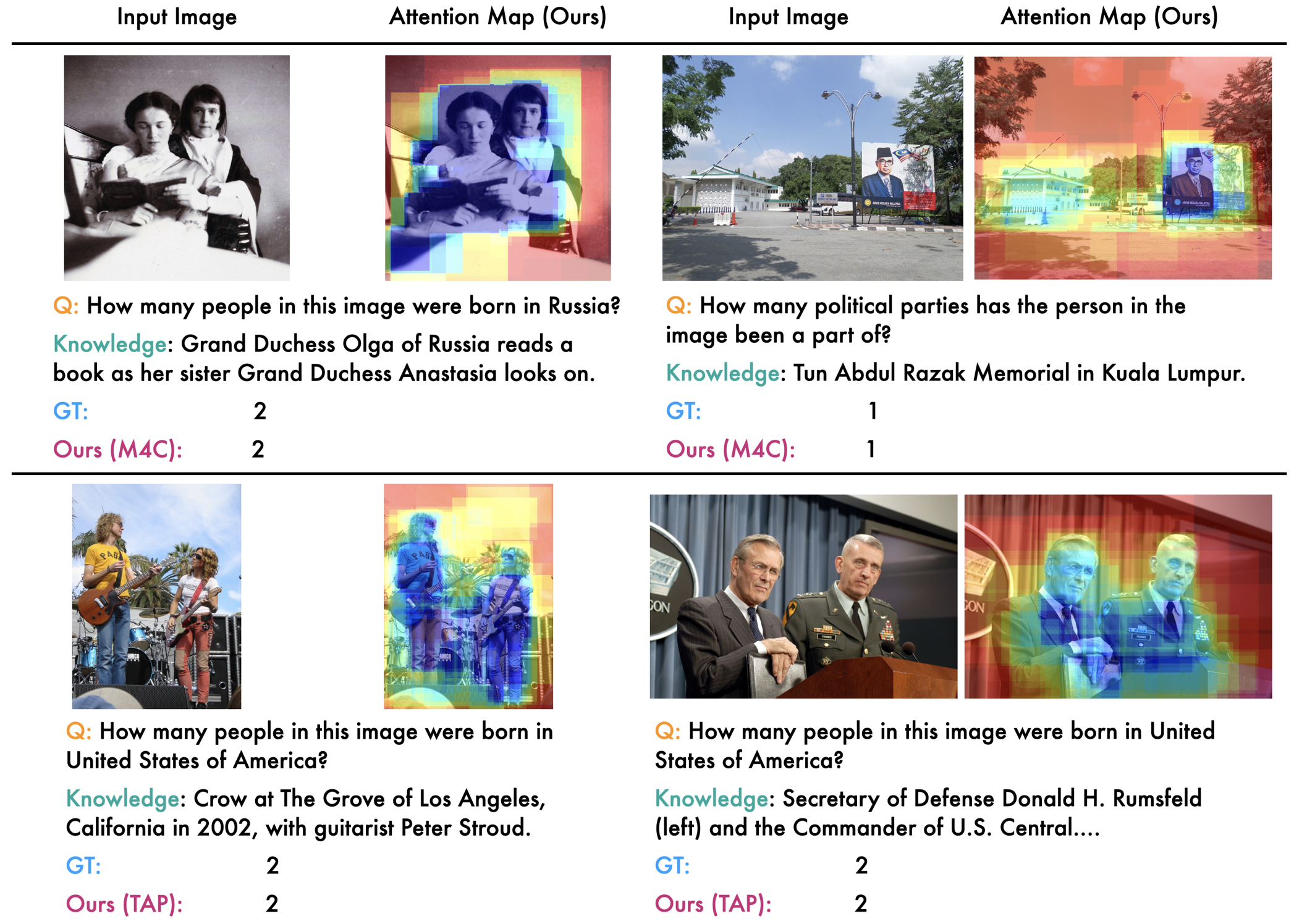}
    \caption{Qualitative results of M4C and TAP: Answering of KVQA questions based on knowledge with attention maps.}
    \label{fig:qualitative_kvqa}
\end{figure*}

\begin{table*}[hbt!]
\centering
\begin{tabular}{l c c r}
\toprule
{\textbf{Method}} & {\textbf{Pre-training data}} & {\textbf{Fine-tuning data}} & {\textbf{Test Acc.}} \\ \midrule
MemNet \cite{kvqa}   &  -  & KVQA  & 50.2    \\ 
UNITER \cite{uniter_kvqa}   & -   & KVQA  & \textbf{69.3}    \\ 
M4C     & -     & -   & 22.89    \\ 
M4C     & -    & KVQA   & 47.38  \\ 
TAP   & \begin{tabular}[c]{@{}c@{}}TextVQA + VQA + STVQA\end{tabular}   & -  & 15.68    \\ 
TAP   & \begin{tabular}[c]{@{}c@{}}TextVQA + VQA + STVQA\end{tabular}  & KVQA    & 47.49    \\ \bottomrule
\end{tabular}
\caption{Evaluation of KVQA \cite{kvqa} test data on text-based VQA models trained on our Union dataset.}
\label{tab:kvqa_quant}
\end{table*}

\subsection{Quantitative results}
\label{subsec:quan_res}

We evaluate the performance of the text-based models trained on our Union dataset and compare it against the state-of-the-art models, namely M4C and TAP which are usually trained on TextVQA and STVQA datasets. The comparative results are shown in Table.~\ref{tab:textvqa_quant} using accuracy as the evaluation metric, although we cannot quantitatively display the reduction in bias.
Among the two main baselines, M4C trained on our Union dataset slightly outperforms the existing M4C trained on TextVQA however, as shown in Sec. \ref{subsec:qual_res}, our model predicts better answers that are unbiased and predicted by looking at the appropriate visual features of the images.
In the case of TAP, our model slightly under-performs compared to the TAP model originally trained on just the TextVQA dataset.
The main reason for the decrease in accuracy in case of TAP or the small increase in accuracy in case of M4C is because the models now rely lesser on the bias to answer the questions. With the reduced bias, using more relevant and appropriate data would assist the model to predict correct answers and thus increase the accuracy.

\subsection{Qualitative results}
\label{subsec:qual_res}

\begin{figure*}[!htb]
    \centering
    \includegraphics[width=1\linewidth]{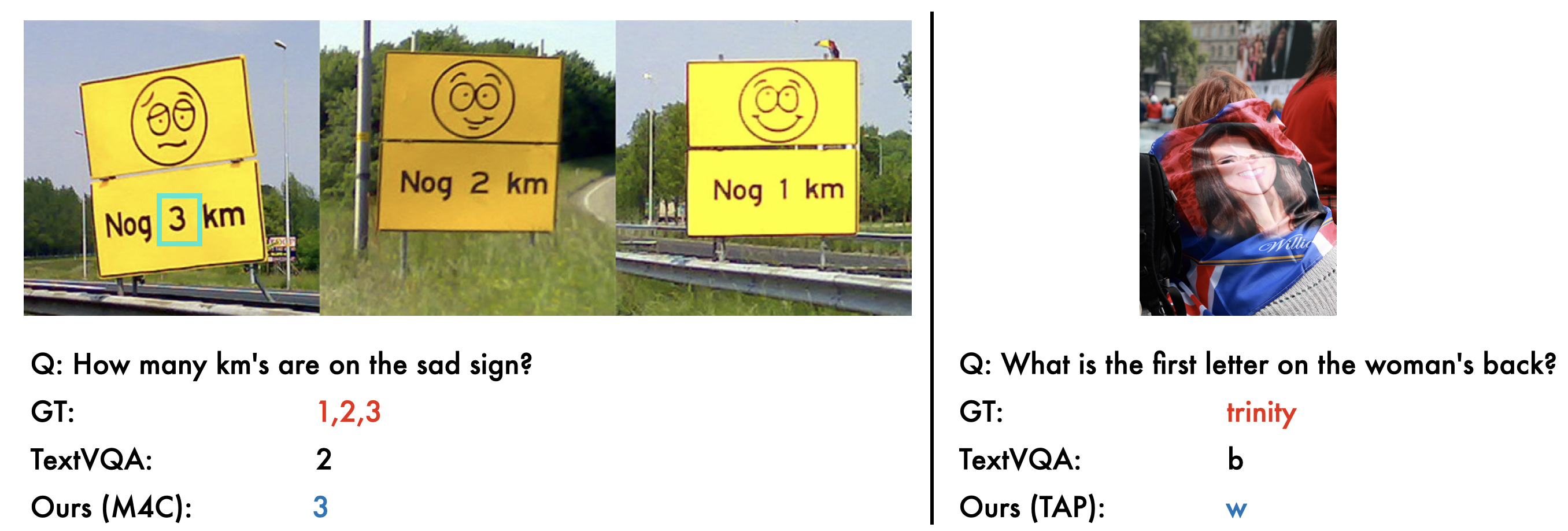}
    \caption{Ground Truth Limitations: In a few cases the dataset contains ambiguous answers such as the first case where the number of km's on the sad sign is asked, but the ground truth answer provided is \textit{1, 2, 3} which is incorrect. }
    \label{fig:negative_samples}
\end{figure*}

We showcase a detailed qualitative analysis of our proposed method. In Fig.~\ref{fig:qualitative_m4c}, we show the predictions of M4C model trained on the union dataset and predictions of M4C model trained only on TextVQA dataset. It can be seen that our method predicts correct answers for the questions that require both visual and textual understanding to answer the question. Example: \textit{``What number is on the middle bike?''}, the answer to this question predicted by our method is \textit{``30''}, whereas the M4C trained only on TextVQA has predicted \textit{``598''}. A model trained only on TextVQA suffers from the bias by the type of question-answer pairs present in the dataset. M4C trained only on TextVQA fails to comprehend the meaning of the word ``middle'', whereas M4C trained by our proposed method can answer the question. This is because of the questions in the VQA dataset, which require the models to look at the image to answer a question, thereby understanding the spatial positions and image features while reasoning over the image to answer the question rather than to just read the text. Fig.~\ref{fig:qualitative_tap} shows the qualitative results for predictions made by TAP on the question-answer pairs in TextVQA dataset that require both visual and text reasoning to obtain the answers. It can be seen that, TAP trained on Union dataset can look as well as read the content in the image, whereas original TAP fails to do so. Attention maps for all the examples shown in Fig.~\ref{fig:qualitative_m4c} and \ref{fig:qualitative_tap} show that the proposed method indeed looks at the image based on the question asked and can also read the required textual content to obtain accurate answers. 

\subsection{Ablation study}
\label{subsec:ablation}

We perform two ablation studies to demonstrate the performance of our model in the case (i) of an external knowledge based VQA dataset named, KVQA - Knowledge aware Visual Question Answering~\cite{kvqa} and (ii) wherein the ground truth of a given question in the TextVQA dataset~\cite{textVQA} is wrong.
It can be seen in the Fig.~\ref{fig:qualitative_kvqa} that the text-based models trained on our Union dataset looks into the appropriate image regions based on the given question and external knowledge provided. 
In Table.~\ref{tab:kvqa_quant}, we can see the performance of our text-based models performing well on such datasets as well while giving results that are comparable to the existing models used for the KVQA dataset~\cite{kvqa}. UNITER~\cite{uniter_kvqa} is one of the VQA models generally used for datasets like VQA. It achieves a state-of-the-art accuracy of 69.3 on the KVQA dataset due to its larger pretraining data. Our models get less accuracy (15.68 and 22.89) when it is tested on it without finetuning on the training dataset. However once it has been finetuned, the model is also able to answer such questions that focuses more on the visual features and it achieves an accuracy of approximately 47.49. 
Further in Fig.~\ref{fig:negative_samples}, we also show the performance of our models trained on the Union dataset predicting correct answers for the questions with wrong ground truth by looking at the image features and predicting answers by reasoning over the images. As shown in Fig.~\ref{fig:negative_samples}(a), \textit{``How many km's are on the sad sign?''} our model can differentiate between parts of the image by localizing to the `sad sign' and predicting the number of km's to be \textit{`3'}. On the other side in Fig.\ref{fig:negative_samples}(b), \textit{``What is the first letter on the woman's back?''} is the question to which our model predicts `w' which is correct when compared with the wrong ground truth to be \textit{``trinity''}.

\section{Conclusion}
In this work, we address the problem of focusing on the text present in the image compared to visual features and proposed a method to focus on visual features along with the text present in the image. We use a Union dataset, a combination of both Text-based VQA and VQA datasets. We evaluate our method on the state-of-the-art models. We show that our method attends to the corresponding visual features while answering a question. The qualitative result of the samples with the wrong ground truth show that our method outperforms the existing state-of-the-art models in terms of reasoning over the image. Our exhaustive quantitative and qualitative analysis suggests that having an unbiased dataset can result in better-comprehending models thereby taking a step towards well-designed VQA models that are capable of reasoning over multiple modalities. With more appropriate and unbiased data, we could achieve better results and answering through proper reasoning. Self-supervised training on various captioning datasets would help in better understanding of the image and can act as a substitute for the lack of proper scene-text data.
\label{sec:conclusions}

\clearpage
{\small
\bibliographystyle{ieee_fullname}
\bibliography{egbib}
}

\end{document}